\documentclass[11pt,a4paper]{article}
\usepackage[hyperref]{acl2021}
\usepackage{times}
\usepackage{latexsym}

\usepackage{microtype}
\usepackage{times}
\usepackage{latexsym}
\usepackage{graphicx}
\usepackage{amsfonts}
\usepackage{url}
\usepackage{subfigure}
\usepackage{array}
\usepackage{multirow}
\usepackage{arydshln}
\usepackage{bm}
\usepackage{color}
\usepackage{xcolor}
\usepackage{amsmath}
\usepackage{hhline}
\usepackage[switch]{lineno}  %

\DeclareMathOperator*{\argmax}{\arg\!\max}

\makeatletter
\newcommand{\thickhline}{%
\noalign {\ifnum 0=`}\fi \hrule height 0.8pt
\futurelet \reserved@a \@xhline
}

\makeatother

\aclfinalcopy 

\title{From Solving a Problem Boldly to Cutting the Gordian Knot: Idiomatic Text Generation}

\author{Jianing Zhou \\
  UIUC \\
  IL, USA \\
  \texttt{\small zjn1746@illinois.edu} \\\And
  Hongyu Gong \\
  Facebook AI \\
  WA, USA \\
  \texttt{\small hygong@fb.com} \\\And
  Srihari Nanniyur \\
  Saint Francis High School \\
  CA, USA \\
  \texttt{\small snanniyur@gmail.com} \\\And
  Suma Bhat \\
   UIUC \\
  IL, USA \\
  \texttt{\small spbhat2@illinois.edu} \\}

\date{}

\begin{document}

\setlength{\textfloatsep}{1pt}
\maketitle
\begin{abstract}
We study a new application for text generation---idiomatic sentence generation---which aims to transfer literal phrases in sentences into their idiomatic counterparts.  Inspired by psycholinguistic theories of idiom use in one's native language, we propose a novel approach for this task, which retrieves the appropriate idiom for a given literal sentence, extracts the span of the sentence to be replaced by the idiom, and generates the idiomatic sentence by using a neural model to combine the retrieved idiom and the remainder of the sentence.  Experiments on a novel dataset created for this task  show that our model is able to effectively transfer literal sentences into idiomatic ones. Furthermore, automatic and human evaluations  show  that for this task, the proposed model outperforms  a series of competitive baseline models for text generation.

\end{abstract}

\section{Introduction}

Idiomatic expressions (IEs) are forms of  figurative constructions  whose 
meaning cannot always be inferred from the meaning of the words in the expression (non-compositional) \cite{nunberg1994idioms} and so they are unlike literal expressions, as shown in Table \ref{tab:dataset}. They   
represent an important aspect of  native-like language production \cite{wray2000formulaic,schmitt2020vocabulary, pawley2014two}. 
 Cognitive and corpus linguistics has found them to evoke stronger affective responses when compared with literal sentences  \cite{nunberg1994idioms,citron2019idiomatic}, 
serving to succinctly  convey abstract and concrete concepts, to  enhancing textual coherence \cite{nunberg1994idioms,drew1998figures,simpson2003corpus,fernando1996idioms,gibbs2007idioms,citron2019idiomatic}.  

At the heart of methods developed for natural language generation (NLG) lies the ability to identify and generate paraphrases. Despite the significant advances in NLG demonstrated by generative pre-trained transformer models (e.g., \cite{radford2019language,lewis2020bart}), including generating fluent and grammatical English,  it is unclear how good these systems are for generating IEs.   
This is important because   the absence of or inappropriately used IEs in generated texts can result in key semantic, pragmatic and discursive elements being lost in generated texts.




Idioms have been a long-standing challenge in computational linguistics due to their phrase-like form, word-like function and non-compositional meaning  \cite{sag2002multiword,Baldwin2010MultiwordE}.
Despite the large number studies to identify, understand and translate IEs \cite{muzny2013automatic,liu2016phrasal,gong2017geometry,fadaee2018examining,shirin2018replacing}, to the best of our knowledge, there has been no study on their automatic generation and this study aims to fill this research gap. 
More specifically,  given  a sentence with a literal meaning (compositional), we  study how neural models can automatically transform it to a sentence with an IE, while preserving the  original content  in the rewritten sentence. A model with this functionality could be incorporated into  existing text generation systems as a post-processing module to enhance a sentence's pragmatic and discursive role. 

Because of the novelty of this task, we created a dataset of parallel literal and idiomatic sentences. We then used the dataset to compare a set of competitive NLG methods from prior art, including a method that we propose, for the task of idiomatic sentence generation. Our proposed approach is a set of modules that are simple to train and is inspired by a key idea from  cognitive linguistic theories of formulaic expression use, which  posit that humans produce IEs by having a mental lexicon of multiword expressions and idioms \cite{jackendoff1995boundaries,gibbs2007idioms,sprenger2019development}. 

One of the challenges for our task is to generate sentences with non-compositional IEs. Using the idea of cognitive linguistic theory in a sentence generation framework, we decide to first retrieve IEs and then incorporate them into an output sentence than directly generating sentences with IEs. We implement our pipeline system using  a retrieval model, an extraction model and a generation model. The retrieval and extraction models  work by  using a pre-built lexicon of idioms to select an idiom that semantically matches a phrase  in the literal sentence  (retrieval), then identifying the span of the phrase in the literal sentence to be replaced by the idiom (extraction). Finally, the generation model uses the chosen idiom in the rewritten sentence and polishes the text. 

Another challenge is to accurately copy the context from the input literal sentence into the output  while also generating the IE to be inserted. For this challenge, we propose a Guided \textsc{CopyNet} as the generation model. Unlike \textsc{CopyNet} which learns whether or not to copy from the input without explicit supervision, Guided \textsc{CopyNet} is explicitly guided to copy or to generate the tokens more accurately by providing the knowledge from the extraction model.
An important assumption here is that only known idioms (available in the lexicon) can be used in the generated sentences. Compared with end-to-end models, our pipeline system has better explainability owing to the different stages and also demonstrates stronger performance as will be discussed in experiments.


Our contributions are summarized below:\\
\noindent (1) We create the first dataset of parallel idiomatic and literal sentences spanning 876 idioms, including details of the syntactic flexibility of a subset of the idioms. \\ 
\noindent (2)  We study how a set of computational approaches based on prior art and our own proposed model can be leveraged for the task of  idiomatic sentence generation;\\
(3) By automatic and manual evaluations we compare the performance of the studied approaches on the task of idiomatic sentence generation. 

\section{Dataset}
\label{sec:dataset}

The dataset that we create consists of two main components: (1) A set of sentences with IEs and their literal counterparts manually created from a list of IEs and example sentences, and (2) a large set of sentences with IEs that occur `in the wild' from a  publicly available corpus, also with manually created literal counterparts. We provide details of each of these sets below. 

\begin{table}[t]
\small
\setlength{\belowcaptionskip}{0.1pt}
    \centering
    \begin{tabular}{m{3.4cm}|m{3.4cm}}
    \thickhline
        \textbf{Literal sentences} & \textbf{Idiomatic sentences}  \\
    \thickhline
        Mr. Sandler {\color{red}{effectively dealt with the difficult situation}} that was strangling the market. & Mr. Sandler effectively {\color{blue}{cut the Gordian knot}} that was strangling the market.\\
        \hline
        The visitors {\color{red}{headed for shelter}} when it started to rain. & The visitors {\color{blue}{ran for cover}} when it started to rain. \\
        \hline
    \thickhline
    \end{tabular}
    \caption{Examples of idiomatic sentences. Idioms are highlighted in blue, and literal paraphrases are in red.}
    \label{tab:dataset}
\end{table}



\subsection{Online-sourced dataset (ONLINE)}
The dataset has two components: (i) a  lexicon of IEs  and their definitions and, (ii) a parallel corpus of sentences with IEs (used in their idiomatic sense)  and their literal counterparts. There were a total of 876 IEs, which were collected from an online resource\footnote{www.theidioms.com. We obtained its authorization for data access.}. We note that almost all IEs have multiple definitions with an average of 4.36 definitions per IE. Some idioms such as ``tick off'' have multiple senses. The annotators labeled the sense of the IEs in the sentences according to the sense information from the same online resource. For each IE, we collected sentences containing the idiom (average of {6.32} sentences per IE), yielding a total of $5537$ idiomatic sentences from the same online resource.

A native English speaker rewrote each idiomatic sentence into a literal sentence, where the IE was replaced with its literal paraphrase.  After rewriting, another annotator with native-level English proficiency checked sentences and corrected any errors to guarantee data quality. 
 We note that each instance (idiomatic and literal) is only one sentence long. Table~\ref{tab:dataset} shows some examples in our corpus. 



\begin{table}[t]
\small
\setlength{\belowcaptionskip}{0.01cm}
    \centering
    \setlength{\tabcolsep}{1mm}
    \begin{tabular}{c|c|c}
    \thickhline
        Statistics & \# of instances & Avg. \# of words \\
        \hline
        Idioms & 876 & 3.4 \\
        \hline
        Definitions & 4548 & 10.4\\
        \hline
        Idiomatic sent & 5537 & 18.6\\
        \hline
        Literal sent & 5537 & 18.1\\
        \thickhline
    \end{tabular}
    \caption{Statistics of our online-sourced parallel corpus.}
    \label{tab:statistics}
\end{table}

\subsection{MAGPIE-sourced dataset (MAGPIE)}

Considering that the IEs and parallel examples collected from the online resources are on a small scale than needed for training neural models, we enlarge the number of parallel instances using the publicly available MAGPIE dataset \cite{haagsma2020magpie}, a collection of sentences with IEs collected from the British National Corpus. We first excluded the sentences with IEs used in a literal sense using the labels provided in the dataset. Then we excluded sentences longer than 30 words to keep them comparable to the manually created dataset and to avoid any long-range dependency challenges for generation. This resulted in 1536 idioms and 17000 idiomatic sentences over 1536 IEs. 


More examples of the corpus are available in the Appendix. We summarize the corpus statistics in Table~\ref{tab:statistics}. Our parallel corpus is publicly available\footnote{https://github.com/zhjjn/PIE.git}. The two datasets serve complementary functions; with the IEs, their definitions and corresponding synthetic examples, ONLINE serves as the lexicon for training the idiom retrieval and span extraction modules of our proposed system. MAGPIE, on the other hand provides idiomatic examples at scale (with naturally occurring IEs) for training the generation model.  
 
 For a subset of the dataset (291 out of 876, used as part of our evaluations in the Experiments), we manually annotated each idiom for the level of lexical rigidity that governs the words of the idioms. This subset was chosen based on the idioms in the original list that were also in WordNet (to ensure that the sense distinctions were accurate). Toward this, we used 3 levels:  Fixed expressions (level 1), semi-fixed expressions (level 2) and syntactically-flexible expressions (level 3) \cite{sag2002multiword}. The definitions  for each level of lexical rigidity are provided in the Appendix. Two researchers annotated the idioms, and reconciled differences  to yield  100\% agreement on the labels.
 
 In addition to creating the parallel corpus, we annotated the literal phrases with BIO labels \cite{ramshaw1999text} to mark the ground truth span of the phrases to be replaced in the input sentences (the phrases colored red in Table~\ref{tab:dataset}). The first word of the span was labeled as ``B'', and the other words within the span were labeled as ``I''. All words outside the span were labeled as ``O''. These labels were  used  to train our proposed model (elaborated later in the Model Section).

\section{Model}
\label{sec:model}
The task of idiomatic sentence generation is to rephrase a given literal sentence into its idiomatic counterpart by using an IE to replace a literal phrase while preserving the original meaning of the sentence. This task can be regarded as paraphrasing only a portion of  the original sentence because we retain the remaining portion intact. 
We use ideas about native speakers accessing a mental lexicon of  formulaic expressions, including IEs  \cite{jackendoff1995boundaries,gibbs2007idioms,sprenger2006lexical,sprenger2019development} to propose a generation model built upon a pipeline of three modules that perform idiom retrieval, span extraction and idiomatic sentence generation. 

\begin{figure}[t]
\setlength{\belowcaptionskip}{1pt}
    \centering
    \includegraphics[width = 0.5\textwidth]{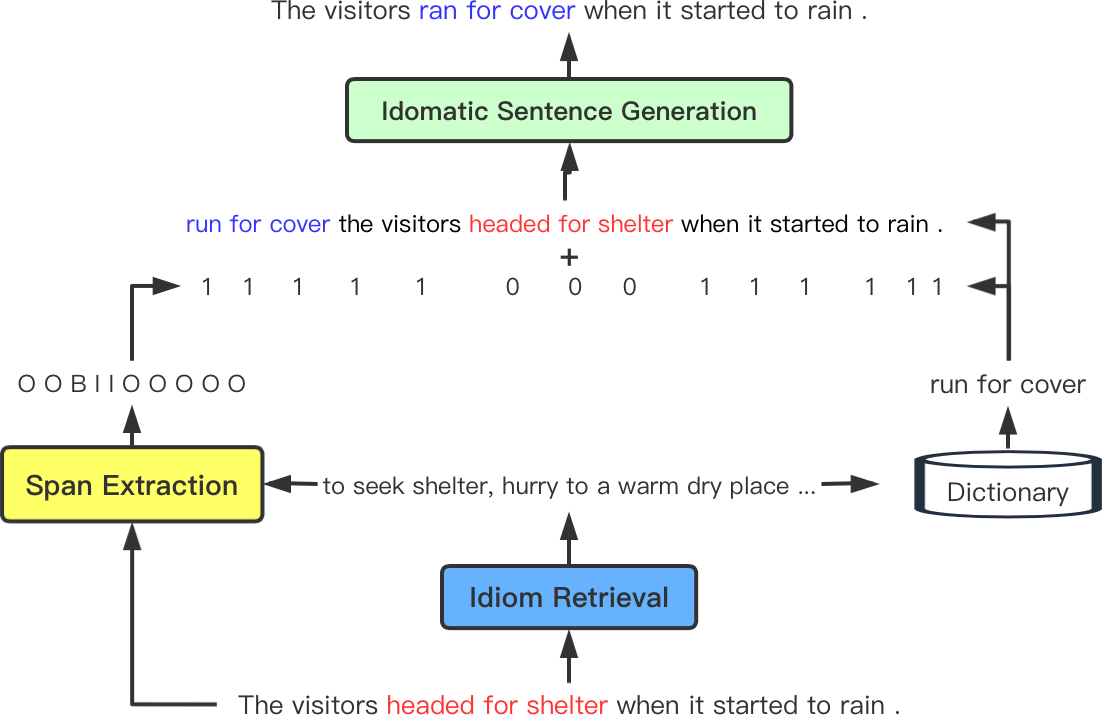}
    \caption{The workflow of the pipeline model for idiomatic sentence generation.}
    \label{fig:pipeline}
\end{figure}

An illustration of the pipeline is shown in  Fig.~\ref{fig:pipeline}. The input literal sentence is ``The visitors headed for shelter when it started to rain .'' The idiom retrieval module, using the available idioms and their definitions,  retrieves an idiom that fits in  this sentence well, which for this example is ``run for cover''. This idiom will then be used in our generated text. Following this, the span extraction module decides the span of the literal sentence to be replaced with the selected idiom. The selected span is ``headed for shelter'', a semantic equivalent of the idiom ``run for cover''.
Lastly, the generation module generates the idiomatic sentence based on the retrieved idiom and the input sentence marked with the selected span. Fig.~\ref{fig:pipeline} shows the generated sentence ``The visitor ran for cover when it started to rain .'', where the selected span is replaced with the retrieved idiom. We will next elaborate upon each module.

\subsection{Idiom Retrieval}

We use the  lexicon with idioms and their definitions created as part of the dataset described in Section~\ref{sec:dataset}. The module for idiom retrieval searches an idiom that  best fits  the given literal sentence. It is built upon a pretrained RoBERTa model \cite{liu2019roberta} and a feed-forward classifier. The RoBERTa model takes as input a sequence of tokens, and generates a contextualized representation for each token as well as the whole sequence. The classifier takes the learned representation and predicts whether an idiom fits in well with the given sentence. Owing to space constraints, the details of the RoBERTa model are provided in the Appendix.





\subsection{Span Extraction}

After selecting the idiom for a given sentence $s$, we need to decide which phrase of the input literal sentence should be replaced by this idiom. The span extraction module extracts the span of the words of the phrase from the input sentence. We use  the ground truth BIO labels marking the span of the phrase in the input sentence (refer to the Dataset section) and cast  the span extraction task  as a sequence labeling problem.

Our span extractor consists of a RoBERTa model and a classifier based on Conditional Random Field \cite{sutton2007dynamic}. The RoBERTa model learns the contextualized representations, which are used by the CRF classifier to label each token in the literal sentence with the B, I, O labels. More details are in the Appendix.



\subsection{Idiomatic Sentence Generation}

\begin{figure*}[t]
    \centering
    \includegraphics[width = \textwidth]{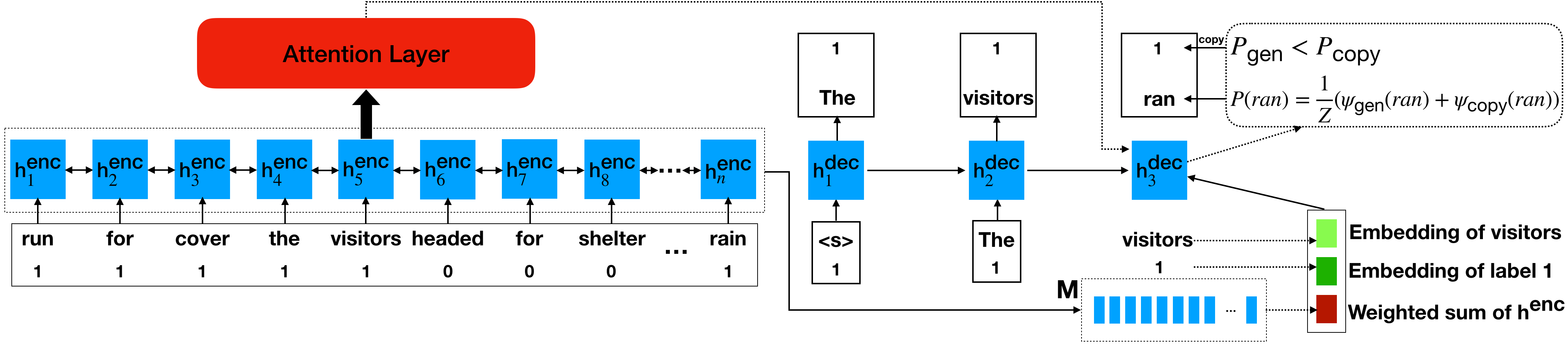}
    \caption{Model Overview of Guided \textsc{CopyNet}}
    \label{fig:copynet}
    \vspace{-5mm}
\end{figure*}

Based on idiom retrieval and span extraction, we obtain the idiom for rewriting and the span of a literal sentence to be replaced by the idiom. Lastly, we propose a Guided \textsc{CopyNet} to generate idiomatic sentences. We provide a detailed description of the generation module next.



\section{Guided \textsc{CopyNet}}
\label{sec:guided_copynet}

The last module of our pipeline is a text generator, which generates idiomatic sentences given the literal sentences and the retrieved idioms. As can be seen in Table-\ref{tab:examples}, a small span of the input is to be replaced with the idiom and the rest of the  input words are to be retained in the generated sentence. To this end, our generation module is inspired by the copy mechanism introduced by \textsc{CopyNet}  \cite{gu2016incorporating}, 
a  text summarization model whose copy mechanism enables the model to copy words directly from input sentences. 


Unlike \textsc{CopyNet}, our generation module does not simply take a literal sentence as input and generate an output sentence. As we will show in the experiments, \textsc{CopyNet} cannot generate a good quality output. 
We propose a Guided \textsc{CopyNet} as the generation module, which is guided by the retrieval module and span extractor. Besides the literal sentence, Guided \textsc{CopyNet} leverages the information of the retrieved idiom and the extracted input span towards generating an output. 


\subsection{Generator Design}

We first give an overview of Guided \textsc{CopyNet}. Similar to \textsc{CopyNet}, it has an encoder-decoder structure as illustrated in Figure~\ref{fig:copynet}. 

\noindent\textbf{Encoder}. The encoder is a Bidirectional Gated Recurrent Unit (BiGRU) model. The input to the encoder is a concatenation of the retrieved idiom and the literal sentence. Each word in the input is represented by a vector which concatenates the word embedding and the copy embedding. The word embeddings are initialized randomly. 

The copy embedding carries the knowledge from the span extraction module. Recall that the span extractor identifies the span to be replaced in a literal sentence, and we can learn which words should be retained and which words should be replaced in the output sentence. As shown in  Fig.~\ref{fig:copynet}, we assign a copy indicator (either 0 or 1) to each word in the input to indicate whether it should be copied to the output or not. The words in the idiom have a copy indicator ``1'' since they will be used in the generated idiomatic sentence. The words in the literal sentence are labeled as ``1'' if they are outside the extracted span, and labeled as ``0'' otherwise. We create $32$-dimension embeddings for the two copy indicators, which are randomly initialized and are tuned during training.

Suppose that word $k$ in the input has word embedding $\mathbf{w}_{k}$ and copy embedding $\mathbf{c}_{k}$. The encoder processes input words sequentially. It takes word $k$ at time $k$, and updates its hidden state vector $\mathbf{h}^{\text{enc}}_{k-1}$ to $\mathbf{h}^{\text{enc}}_{k}$ as below.
\begin{align*}
    \mathbf{h}^{\text{enc}}_{k} = \text{BiGRU}(\mathbf{h}^{\text{enc}}_{k-1}, \mathbf{w}_{k}\oplus\mathbf{c}_{k}).
\end{align*}
The encoder generates a sequence of hidden state vectors $ \textbf{M} = \{\mathbf{h}^{\text{enc}}_{1},\ldots, \mathbf{h}^{\text{enc}}_{k}, \ldots\}$ corresponding to each word.

\noindent\textbf{Decoder}. The decoder is built using GRU and an attention layer. It takes an input vector, updates its hidden state vector  and generates one word at a time. Suppose that the word generated at time $t-1$ is $y_{t-1}$. At time $t$, the input vector is a combination of the embedding $\mathbf{w}_{t-1}$ of last output word $y_{t-1}$, the label embedding $\mathbf{l}_{t-1}$ of word $y_{t-1}$, the context embedding $\mathbf{s}_{t-1}$ and a weighted sum of hidden states in $\textbf{M}$ denoted by $\bm{\xi}_{t-1}$. Next we will discuss each of these embeddings.

The word embeddings are the semantic vectors of the output words and are initialized randomly. The label embedding in the decoder is similar to the copy embedding in encoder. 
For the token generated at time $t-1$, we know whether it is copied from input or not because if copied from the input it is labeled as 1 and as 0 otherwise. The embeddings for the label set $\{0,1\}$  are randomly initialized and tuned during training. 

As for the context embedding, we use the attention layer in \cite{gu2016incorporating} to assign attention to the input words, and derive the context embeddings as a  sum of the encoder's hidden state vectors weighted by the attention. The embedding $\bm{\xi}_{t-1}$ is a weighted sum of hidden states $\textbf{M}$ from encoder:
 \begin{align*}
 \setlength{\belowdisplayskip}{0.1cm}
     \begin{split}
         &\bm{\xi}_{t-1} = \sum^{n}_{k = 1} \alpha_{t k}\textbf{h}^{\text{enc}}_{k}
     \end{split}
 \end{align*}
where the weight coefficient is $\alpha_{t k} = \frac{1}{K} \psi_{\text{copy}}(y_{t-1}=x_{k})$, and the normalization term is $K = \sum_{k':x_{k'}=y_{t-1}} \psi_{\text{copy}}(y_{t-1}=x_{k'})$.

Therefore, the decoder updates its hidden vector $\mathbf{h}^{\text{dec}}_{t}$ at time $t$ as below.
\begin{align*}
   \mathbf{h}^{\text{dec}}_{t} = \text{GRU}(\mathbf{h}^{\text{dec}}_{t-1}, \mathbf{w}_{t-1}\oplus\mathbf{l}_{t-1}\oplus\mathbf{s}_{t-1}\oplus\bm{\xi}_{t-1}). 
\end{align*}

\noindent\textbf{Copy Mechanism}. The word in the output sentence is predicted based on the decoder's hidden state vectors $\{\mathbf{h}^{\text{dec}}_{t}\}_{t}$. With the copy mechanism of \textsc{CopyNet}, the words are predicted in either copy-mode or generate-mode. In copy-mode, the score $\psi_{\text{copy}}(y_{t}=x_{j})$ for copying input word $x_{j}$ as output word $y_{t}$ is estimated as
\begin{align*}
    \psi_{\text{copy}}(y_{t}=x_{j})
    =\sigma({\mathbf{h}^{\text{enc}}_{j}}^{T}\mathbf{U})\mathbf{h}^{\text{dec}}_{t},
\end{align*}
where $\sigma$ is $tanh$ activation function, and $\mathbf{U}$ is a parameter.

In generation mode, the score $\psi_{\text{gen}}$ for generating word $v_{k}$ in the output vocabulary is as
\begin{align*}
   \psi_{\text{gen}}(y_{t}=v_{k}) = \mathbf{v}_{k}^{T}\mathbf{W}\mathbf{h}^{\text{dec}}_{t}, 
\end{align*}
where $\mathbf{v}_{k}$ is a one-hot vector for word $v_{k}$, and matrix $\mathbf{W}$ is a model parameter.

We predict the probability that $y_{t}$ is word $w$:
\begin{align*}
p(y_{t}=w) = \frac{1}{Z}(\psi_{\text{copy}}(y_{t}=w) + \psi_{\text{gen}}(y_{t}=w)),
\end{align*}
where $Z$ is a normalization factor and is the sum of all scores $\psi_{\text{copy}}(\cdot)$ and $\psi_{\text{gen}}(\cdot)$.

\section{Experiments}

In this section, we empirically evaluate our pipeline model for idiomatic sentence generation, and compare it with  text generation models that have been found to perform well in other text generation tasks. To gain insights into the performance of each module in the pipeline, we further assess the idiom retrieval, the span extractor and the Guided \textsc{CopyNet} modules individually.
 The subset of $291$ idioms 
in Section \ref{sec:dataset} is the annotated subset of idioms for evaluation. 
For each idiom in the annotated subset, we randomly selected one parallel sentence pair into the validation set, one sentence pair into the test set and all the other sentence pairs into the training set. If there were only two sentence pairs for one idiom, they were randomly split between the training and test set. For each idiom not in the annotated subset, all the sentence pairs were included in the training set. Without augmentation of MAGPIE dataset, there were 4997, 249 and 291 sentence pairs in the training, validation and test sets respectively. After augmentation, there were 21997, 249 and 291 sentence pairs in each set.


\subsection{Baselines}

Considering that there has been no prior work on the new task of idiomatic sentence generation, we include three state-of-the-art text generation models as baselines for this task. \\
\noindent(1) Seq2Seq model \cite{luong17}: an encoder-decoder model built on Long Short Term Memory (LSTM) used in machine translation; \\
\noindent(2) Transformer \cite{vaswani2017attention}: a deep neural network with self-attention mechanism; \\
\noindent(3) Copy-enriched Seq2Seq model \cite{jhamtani2017shakespearizing}: an LSTM-based Seq2Seq model which is able to copy directly from inputs; \\
\noindent(4) Copy-enriched Transformer model\footnote{https://github.com/lipiji/TranSummar}: a Transformer-based Seq2Seq model which is able to copy directly from inputs.


Details for training of these baselines are provided in the Appendix.

The generator in this task needs to appropriately use the retrieved idioms in transferred sentences. \cite{li2018delete} used a simple rule-based method to generate good sentences by directly replacing the extracted input span with the retrieved idiom. To gain insights about the benefits of Guided \textsc{CopyNet} for this task, we also compare it with this rule-based approach and the original \textsc{CopyNet}.

\subsection{Experimental Setup}

\begin{table*}[t]
\small
    \centering
    \begin{tabular}{ccccccc}
    \thickhline
        Model & BLEU & ROUGE-1 & ROUGE-2 & ROUGE-L & METEOR & Perplexity\\
        \thickhline
        Seq2Seq
        & 55.51 & 70.69 & 54.42 & 71.07 & 67.48 & 7.69\\

        Transformer & 60.29 & 63.15 & 53.35 & 64.55 & 70.55 & 7.71\\

        Seq2Seq with copy & 55.86 & 70.81 & 56.72 & 73.70 & 78.75 & 7.24\\
        
        Transformer with copy & 61.87 & 68.80 & 57.44 & 69.83 & 78.21 & 7.27\\
        \hline
        Ours - original \textsc{CopyNet} & 61.75 & 75.81 & 60.46 & 76.52 & 76.28 & 5.76\\
        \cdashline{1-7}[1.5pt/2pt]
        Ours -Rule based & 62.13 & 76.74 & 64.45 & 77.89 & 73.62 & 5.55\\
        \cdashline{1-7}[1.5pt/2pt]
        Ours - Guided \textsc{CopyNet} & \textbf{65.67} & \textbf{79.56} & \textbf{66.87} & \textbf{79.97} & \textbf{79.74} & \textbf{5.01}\\
        \thickhline
    \end{tabular}
    \caption{The overall performance of baselines and our model based on the augmented training set for generation.}
    \label{tab:overall2}
\end{table*}

\begin{table*}[t]
\small
    \centering
    \begin{tabular}{ccccccc}
    \thickhline
        Model & BLEU & ROUGE-1 & ROUGE-2 & ROUGE-L & METEOR & Perplexity\\
        \thickhline
        Seq2Seq
        & 41.92 & 60.64 & 40.31 & 60.79 & 56.53 & 15.33\\

        Transformer & 39.82 & 51.31 & 36.87 & 51.62 & 58.92 & 15.71\\

        Seq2Seq with copy & 46.66 & 61.94 & 46.46 & 65.48 & 66.44 & 15.13\\
        
        Transformer with copy & 51.50 & 62.25 & 48.71 & 65.56 & 68.42 & 13.05\\
        \hline
        Ours - original \textsc{CopyNet} & 61.39 & 71.81 & 58.71 & 71.90 & 74.2 & 7.86\\
        \cdashline{1-7}[1.5pt/2pt]
        Ours -Rule based & 62.13 & 76.74 & 64.45 & 77.89 & 73.62 & \textbf{5.55}\\
        \cdashline{1-7}[1.5pt/2pt]
        Ours - Guided \textsc{CopyNet} & \textbf{62.34} & \textbf{77.24} & \textbf{64.70} & \textbf{78.82} & \textbf{74.37} & 7.01\\
        \thickhline
    \end{tabular}
    \caption{The overall performance of baselines and our model based on the training set without augmentation.}
    \label{tab:overall3}
    \vspace{-5mm}
\end{table*}

The three modules in our pipeline model described in Section \ref{sec:model} were trained independently. During the evaluation, these modules were stacked sequentially, and the retrieved idioms from the retrieval module were sent to the span extractor for span extraction. Lastly, Guided \textsc{CopyNet} generated sentences based on the outputs of these two modules. 
The detailed experimental setup is provided in the Appendix due to space constrains.


\subsection{Automatic Evaluation}

For the overall evaluation, we evaluated the quality of the generated idiomatic sentence---the output of our pipeline model. Rouge-1, Rouge-2, Rouge-L \cite{lin-2004-rouge}, BLEU \cite{papineni2002bleu} and METEOR \cite{lavie2007meteor} were used as evaluation measures to compare the similarity between the generated sentences and the references. To measure linguistic quality, we fine tuned OpenAI GPT-2 on the target sentences using the same dataset split used for generation and use it to measure the perplexity of the generated sentences.

We also evaluated each module in the pipeline. For the evaluation of idiom retrieval, we calculated the accuracy of idioms retrieved for literal sentences.
As for span extraction, we consider words with predicted labels of ``B'' or ``I'' to be in the predicted span.
Borrowing ideas from the evaluation of machine reading comprehension \cite{rajpurkar2016squad} we report the macro F1 score of the predicted spans in comparison with the ground truth. For a literal sentence $s$, suppose that the predicted span has $L^{s}_{p}$ words, the target span  $L^{s}_{t}$words, and the two spans  $L^{s}_{c}$ words in common. 

The precision $P_{s}$ and the recall $R_{s}$ for sentence $s$, and the macro F1 score $F1$ are defined as 
\begin{equation*}
\small
\begin{split}
    P_{s} &= \frac{L^{s}_{c}}{L^{s}_{p}}; R_{s} = \frac{L^{s}_{c}}{L^{s}_{t}}; F1 = \frac{1}{|S|}\sum\limits_{s\in S}\frac{2P_{s}R_{s}}{P_{s}+R_{s}}.
\end{split}
\end{equation*}

\subsection{Human Evaluation}
We also included human evaluation of the generated sentences to be complementary to the automatic evaluation metrics. 
We randomly sampled $60$ literal sentences and the corresponding outputs of our system as well as all baselines. Human annotations were collected with respect to content, style and fluency of these generated sentences based on the following criteria.\\
\noindent (1) \textbf{Context preservation} measures how well the context is preserved in the output. \\
\noindent (2) \textbf{Idiom inclusion} checks whether the correct idiom is used in the output. \\
\noindent (3) \textbf{Fluency} evaluates the fluency and readability of the output sentence including how appropriately the verb tense, noun and pronoun forms are used.
\noindent (4) \textbf{Overall meaning} evaluates the overall quality of the output sentence.

For each output sentence, two native speakers were asked to rate it on a scale from $1$ to $6$ in terms of the context preservation, fluency and overall meaning. As for the idiom inclusion, they were asked to rate it on a scale from $1$ to $3$. Higher scores indicate better quality. Details of the evaluation scales are provided in Appendix.

\section{Results}
We evaluate the overall performance of the pipeline in the following. The performance of each module in the pipeline model is also evaluated individually, which is provided in the Appendix. 
\subsection{Overall Results and Analysis}

\begin{table}[t]
\small
    \centering
    \begin{tabular}{ccccc}
    \thickhline
        Model & Content & Idiom & Fluency & O \\
        \thickhline
        Seq2Seq & 1.3 & 1.1 & 1.1 & 1.7 \\
        Seq2Seq - copy & 3.8 & 1.6 & 2.1 & 3.5 \\
        Transformer & 4.2 & 1.3 & 3.3 & 3.4 \\
        Transformer - copy & 5.4 & 1.2 & 4.6 & 3.9 \\
        Ours - original & 5.4 & 1.4 & 3.8 & 4.1 \\
        Ours - Rule based & \textbf{5.6} & \textbf{1.7} & 4.2 & 4.2 \\
        Ours - Guided & \textbf{5.6} & \textbf{1.7} & \textbf{5.1} & \textbf{4.5} \\
        \thickhline
    \end{tabular}
    \caption{Human evaluation results. O denotes Overall meaning scores.}
    \label{tab:manual}
\end{table}

We report in Table \ref{tab:overall2} the final results of the idiomatic sentences generated by our pipeline and three strong baselines. In particular, we tried both original \textsc{CopyNet} and Guided \textsc{CopyNet} as the generation module in our pipeline. Their performance is also included in Table \ref{tab:overall2}. As can be seen, the pipeline with Guided \textsc{CopyNet} achieves the best performance in all evaluation metrics. Its improvement over Seq2Seq model with copy mechanism is $9.81$ in BLEU and $10.15$ in ROUGE-2.


\begin{table*}[t]
\small
    \centering
    \begin{tabular}{c m{11.8cm}}
    \thickhline
    literal sentence & She woke up early in the morning and started {\color{red}{thinking deeply}}  {\color{red}{over}} things. \\
    \hline
    Reference &She woke up early in the morning and started {\color{blue}{mulling}} things {\color{blue}{over}}. \\
    \hline
    Seq2Seq & She woke up early in the morning and started thinking sleep things over the button. \\
    Transformer & she woke up early in the morning and started {\color{red}{thinking deeply}} bits . \\
    Seq2Seq with copy & she unk up early in the morning and started unk things over it \\
    Transformer with copy & she woke up early in the morning and started thinking head over . \\
    Ours - original \textsc{CopyNet} & She woke up early in the {\color{blue}{mull}} morning and started started left. \\
    Ours -Rule based & she woke up early in the morning and started {\color{blue}{mull over}} over \\
    Ours - Guided \textsc{CopyNet} & She woke up early in the morning and started {\color{blue}{mulling}} things {\color{blue}{over}} \\
    \thickhline
    \end{tabular}
    \caption{A sample of generated idiomatic sentences. The words in red color are the words should be deleted and the words in blue color are corresponding idiom.}
    \label{tab:examples}
    \vspace{-5mm}
\end{table*}

\begin{table}[t]
\small
    \centering
    \begin{tabular}{c|c|c|c}
    \thickhline
        Idiom type & $\%$ & Rule-based & Guided \textsc{CopyNet} \\
        \thickhline
        Fixed & 36.39 & 65.16 & 64.42 \\
        Semi-fixed & 45.94 & 64.39 & 67.48\\
        Flexible & 17.67 & 57.39 & 61.26\\
        \thickhline
    \end{tabular}
    \caption{BLEU of different types of idioms.} 
    \label{tab:fixedness}
\end{table}

We further evaluated the effect of Guided \textsc{CopyNet} by comparing the performance of the pipeline with Guided \textsc{CopyNet} and that with original \textsc{CopyNet}.  
When the original \textsc{CopyNet} was used as the generation module in the pipeline, its input was the concatenation of the retrieved idiom and the literal sentence with the extracted span removed. We note that in comparison with the original \textsc{CopyNet}, the Guided \textsc{CopyNet} achieves a gain of $4.92$ in BLEU, $6.41$ in ROUGE-2.

We also evaluated the effect of augmentation by the MAGPIE dataset. Table \ref{tab:overall3} showed the performance of the baselines and our model based on the training set without augmentation. Compared with the results in Table \ref{tab:overall2}, we can see that all the models benefited from data augmentation.

Table \ref{tab:manual} shows the human evaluation results. For content preservation and idiom inclusion, Guided \textsc{CopyNet} performs competitively with the Rule-based method. But for fluency, we can see that Guided \textsc{CopyNet} has a better performance than Rule-based method. Overall, our Guided \textsc{CopyNet} achieves the best performance.
Table \ref{tab:examples} shows an example of a literal sentence together with the generated outputs by different models. We see that our pipeline is able to extract the correct idiom and use it appropriately in the sentence. More examples are provided in the Appendix.

For each idiom category (by lexical rigidity), we report the performance of both methods in Table~\ref{tab:fixedness}. We see that the rule based method fares slightly better than Guided \textsc{CopyNet} for fixed idioms. As for the other idiom categories which require syntactic changes, Guided \textsc{CopyNet} outperforms the rule based method since it is able to modify the idioms  based on the context.

\subsection{Analysis}

\noindent \textbf{Our Model}. For our model, the overall performance is restricted by the first two stages because of the nature of the pipeline model. 
The errors of our retrieval module can be seen when multiple idioms have similar meanings (which is technically not an error). For example, given the literal sentence ``Who got you annoyed'', the correct idiom in the dataset was ``get one's goat''. Instead, our retrieval model retrieved the idiom ``wind up'', because ``wind up'' has a similar meaning to ``get one's goat''. Additionally, we observe that the retrieval module selected wrong idioms because of the non-compositionality of idioms as shown in Table \ref{tab:examples2} in the Appendix. Instead of retrieving the correct idiom ``build castles in the air'', our model wrongly retrieved  ``bread and butter'', whose definition is more similar to ``having a lot of money'' appeared in the input. Without sufficient contextual information, the model was unable to accurately match the idioms with the phrases in a few of the sentences. However, for the idioms that are correctly retrieved, our pipeline model has shown the ability to correctly delete the extracted span and insert the retrieved idiom into the rest of the source sentence with the appropriate morphological modifications when necessary.

\noindent \textbf{Baseline Models}. For the baseline models, we  chose some of the more commonly used state-of-the-art end-to-end text generation models, which may not be the optimal choice because of the high similarity between the input and output in the training set. 
We notice that  the baseline models attempt to just copy the input into the output highlighting t he difficulty for these basic models to learn the mapping between the literal phrase and the corresponding idiom. To clarify this, we used our pipeline model with \textsc{CopyNet} as a comparison to see if the mapping between the literal phrase and corresponding idiom can be first learned by the idiom retrieval module and the span extraction module, yielding a performance improvement in \textsc{CopyNet}. Taking ``She woke up early in the morning and started thinking deeply over things.'' in Table \ref{tab:examples} as an example. All the baseline models mainly focus on copying the input into the output, shown by how  ``thinking deeply over'' or parts of it are seen in the outputs of the baseline models.

\section{Related Work}

In this work, we propose a new task of idiomatic text generation, 
which is naturally  connected to two streams of text generation tasks---paraphrasing and style transfer. We will discuss their similarities as well as their differences to the task in this paper.



\noindent \textbf{Paraphrasing}. This task is to rewrite a given sentence and preserve the original meaning. Idiomatic text generation can be considered as constrained paraphrasing with the constraint that idiomatic expressions should be used in outputs. Seq2Seq models have been successfully applied to paraphrasing \citet{prakash2016neural,gupta2018deep,iyyer2018adversarial,yang2019end}. 
Besides the end-to-end models, a template-based pipeline model was proposed to divide paraphrase generation into template extraction, template transforming and template filling \cite{gu2019extract}.


\noindent \textbf{Style Transfer}. The task of style transfer is to rewrite sentences in a way that the text is changed to a target style (e.g., the writing style of specific authors). Style transfer may need to completely change input sentences, while idiomatic rewriting usually retains a large portion of inputs. Supervised approaches include the Pointer-Generator model \cite{jhamtani2017shakespearizing}. Unsupervised models include cross-aligned auto-encoder \cite{hu2017toward}, VAE  \cite{hu2017toward}, Generative Adversarial Network \cite{zeng2020style} and Denoising auto-encoding \cite{subramanian2018multiple}.  Reinforcement learning has been studied to better meet the constraints of style transfer \cite{xu2018unpaired,gong2019reinforcement}. Pipeline models such as the deletion-retrieval-generation pipeline have shown good performance in style transfer  \cite{li2018delete,sudhakar2019transforming}.

\section{Conclusion and Future Work}

In this paper, we proposed a new task for text generation---that of idiomatic sentence generation. We construct the first parallel corpus for this task, and propose a novel pipeline to solve the task via idiom retrieval, span extraction and text generation in sequence. 
Experimental results comparing different text generation methods showed the gains in performance of the proposed pipeline model relative to the competitive systems in text generation by a large margin. 

For future studies, exploring other architectures and pre-trained models is one concrete direction. A second direction of importance would be to account for the prevalence of the idioms as they occur as part of the evaluation. 

\bibliographystyle{acl_natbib}
\bibliography{acl2021}

\clearpage
\appendix

\section{Appendix}
\label{sec:appendix}



\subsection{Parallel Corpus of literal and Idiomatic Sentences}
\label{sec:corpus_app}

\begin{table}[h]
\small
\setlength{\belowcaptionskip}{0.1pt}
    \centering
    \begin{tabular}{m{3.4cm}|m{3.4cm}}
    \thickhline
        \textbf{literal sentences} & \textbf{Idiomatic sentences}  \\
    \thickhline
        To {\color{red}{fully exemplify}} her beauty, you should photograph her in nature; not in a studio & To {\color{blue}{do justice to}} her beauty, you should photograph her in nature; not in a studio. \\
        \hline
        Why don't you just {\color{red}{be quiet}} because I am tired of being nagged all morning? &
        Why don't you just {\color{blue}{zip your lip}} because I am tired of being nagged all morning? \\
        \hline
        His service is awfully expensive but there is none other like him, so I have to {\color{red}{give him the credit he deserves}}. & His service is awfully expensive but there is none other like him so I have to {\color{blue}{give the devil his due}}.\\
    \thickhline
    \end{tabular}
    \caption{Some examples of idiomatic sentences. Idiomatic expressions are highlighted in blue, and their literal paraphrases are in red.}
    \label{tab:dataset2}
\end{table}

In Table \ref{tab:dataset2}, we give more examples of literal and idiomatic sentences from the parallel corpus.

\begin{table}[h]
\tiny
\setlength{\belowcaptionskip}{0.01cm}
    \centering
    \begin{tabular}{c| m{1.7cm} m{1.7cm} m{1.7cm}}
    \thickhline
        Idiom & Definition & Idiomatic example & literal form \\
        \cline{1-4}
        \multirow{3}[19]{0.3in}{find one's feet} & become familiar with a new situation & I'm new to this city, so I'm still finding my feet. & I'm new to this city, so I'm still finding my feet. \\
         \cline{2-4}
         & become confident in what you are doing & It was only after doing many small shows that he finally found his feet as a singer. & It was only after doing many small shows that he finally found his foundation as a singer. \\
         \cline{2-4}
         & become used to a new situation or experience & We have this orientation programme that helps new employees find their feet in the organisation. & We have this orientation programme that helps new employees get accustomed to the organisation. \\
        \thickhline
    \end{tabular}
    \caption{Example of data in our dataset. Simple sentence is the source sentence in our task and Idiomatic sentence is the target sentence.}
    \label{tab:ex2}
\end{table}

As mentioned before, one idiom might have multiple definitions in our parallel corpus. Table \ref{tab:ex2} gives an example of the idiom ``find one's feet'' as well as its three definitions.

\subsection{Definition for Lexical Rigidity Levels}

We relied on the following definitions for each lexical rigidity level as available in \cite{sag2002multiword}.  Fixed expressions are fully lexicalized and undergo neither morphosyntactic variation nor internal modification (e.g., \textit{in short}). Semi-fixed idioms adhere to strict constraints on word order and composition, but undergo some degree of lexical variation (e.g., in the form of inflection, variation in reflexive form, and determiner selection). This makes it possible to treat the idiom as a word complex with a single part of speech. An example is  \textit{turn out} because of the permissible lexical variations on the verb (\textit{turned out, turning out, turns out}). Finally,  syntactically-flexible idioms are those that exhibit a much wider range of syntactic variability.  Idioms of this kind are  decomposable idioms (e.g. \textit{beef up}), which can be varied in their use both with respect to the morphological variants (\textit{beefed up}), but also by inserting words into the idiom (\textit{beef myself up} or \textit{beef his image up}).

\subsection{Details for Baselines}

For baseline models, the dimension of hidden state vectors was set to $256$ and the dimension of the word embeddings  to $128$. The batch size and base learning rates were set to $32$ and $1e-3$. These baselines were trained with the parallel sentence pairs, i.e., they take literal sentences as input and generate the corresponding idiomatic sentences.

\subsection{Details for Idiom Retrieval}
\label{sec:idiom_retrieval}

Suppose that we have an input literal sentence $s$, and an idiom $i$.
The retrieval module prepares a token sequence by concatenating a special token ``[CLS]'', the input literal sentence, the idiom and its definition. The token ``[CLS]'' is added to the beginning of the sequence in order to comply with the input format of RoBERTa. This sequence is fed to RoBERTa, and we derive the sequence embedding $\mathbf{h}^{s}_{\text{ret}}(i)$ from the learned representation of each token in the sequence by adding them together.

The feed-forward classifier takes the sequence embedding and outputs a retrieval score $r^{s}_{\text{ret}}(i)$ to measure how well the idiom $i$ matches sentence $s$.
\begin{align*}
\setlength{\abovedisplayskip}{3pt}
\setlength{\belowdisplayskip}{3pt}
    r^{s}_{\text{ret}}(i) = \mathbf{W}_{\text{ret}}\mathbf{h}^{s}_{\text{ret}}(i)+\mathbf{b}_{\text{ret}},
\end{align*}
where the weight matrix $\mathbf{W}_{\text{ret}}$ and the bias vector $\mathbf{b}_{\text{ret}}$ are parameters of the classifier.

\textbf{Training}. An input instance to the retrieval module was a sentence-idiom pair. An instance was considered as a positive instance and labeled as ``1'', if the idiom was used to rewrite the literal sentence in the parallel dataset. For each positive instance, we also created negative instances with the same literal sentence by randomly sampling $100$ idioms that did not fit with the sentence. These negative instances were labeled as ``0''. The retrieval module was trained with the cross-entropy loss to classify the label of a sentence-idiom pair. Parameters were tuned for both RoBERTa and the classifier in the retrieval module.

\textbf{Test}. Given a literal sentence $s$ during testing, we created an input instance for every idiom $i$ in the dictionary. The retrieval module scores each pair $(s,i)$. The idiom $i^{*}$ with the highest score is then selected for the literal sentence, i.e., $i^{*}=\argmax\limits_{i}r^{s}_{\text{ret}}(i)$.

\subsection{Details for Span Extraction}
\label{sec:span_extraction}

Since the span to be replaced in the literal sentence is semantically similar to the definition of the idiom , we again prepare the input sequence of the span extraction module by concatenating the literal sentence and definition of the idiom. Suppose that the embedding of token $j$ in sentence $s$ learned by the RoBERTa model is $\mathbf{h}^{s}_{\text{ext}}(j)$. A CRF classifier jointly predicts the likelihood $\mathbf{p}^{s}_{\text{ext}}(j)$ over the label set \{B, I, O\} for each token $j$ in the sentence $s$. Suppose that sentence $s$ has $n$ tokens.
\begin{align*}
\setlength{\abovedisplayskip}{-1pt}
\setlength{\belowdisplayskip}{-1pt}
  \mathbf{p}^{s}_{\text{ext}}(1), \ldots, \mathbf{p}^{s}_{\text{ext}}(n) = \text{CRF}(\mathbf{h}^{s}_{\text{ext}}(1), \ldots, \mathbf{h}^{s}_{\text{ext}}(n)),
\end{align*}
where $\text{CRF}(\cdot)$ is the CRF-based sequence classifier.

\textbf{Training}. Both RoBERTa and the CRF classifier in the span extractor are trained using a weighted cross-entropy loss. The weighted loss is adopted to mitigate the imbalanced distribution of labels, since the number of label ``O'' is much larger than that of other labels. The weight is set as 0.48 for the labels "B" and  "I" and 0.04 for the others.

\textbf{Test}. The span extractor outputs labels with the highest likelihood for all tokens in the literal sentence. The tokens with the labels ``B'' or ``I'' correspond to the span we want to replace.

\subsection{Training and Testing For Guided \textsc{CopyNet}}
 
\noindent \textbf{Training}. Given the idiomatic sentence as the target output, the Guided \textsc{CopyNet} is trained to maximize the likelihood of the target tokens in the reference. The word label, which indicates whether the word is copied from the input by the decoder, is known from the reference sentence.

\noindent\textbf{Testing}. When testing, the generation module takes the concatenation of an idiom and a literal sentence as input, and outputs the sentence which has the highest likelihood. We note that the word label used in the decoder is not known since we do not have the reference sentence during testing. We estimate the copy likelihood  $p_{\text{copy}}=\frac{1}{Z}\sum\limits_{w}\psi_{\text{copy}}(y_{t}=w)$, and the generation likelihood $p_{\text{gen}}=\frac{1}{Z}\sum\limits_{w}\psi_{\text{gen}}(y_{t}=w)$. The word label is set as $1$ if $p_{\text{copy}}>p_{\text{gen}}$, and  $0$ otherwise.

\subsection{Experimental Setup}
\noindent \textbf{Idiom Retrieval Module}. We use a RoBERTa-base model with $12$ layers in the retrieval module. The retrieval module takes a pair of a literal sentence and an idiom, and predicts whether the idiom could be used to rewrite the sentence. 
We use AdamW \cite{loshchilov2018fixing} as the optimizer and set the learning rate as $1e-5$.


\noindent \textbf{Span Extractor}. The extraction module also uses a RoBERTa-base model with $12$ layers. The data used to train the span extractor is a sentence with its words labeled as ``B'', ``I'', or ``O'', where ``B'' and ``I'' indicate that the corresponding words will be replaced. 
We use AdamW as our optimizer and set the learning rate as $1e-5$.


\noindent \textbf{Guided \textsc{CopyNet}}. The encoder and decoder were both one-layer GRUs in Guided \textsc{CopyNet}. We set the dimension of their hidden state vectors as $256$. The dimension of the word embeddings was set as $128$ and the dimension of copy embedding was set as $32$. The batch size and base learning rates were set to $32$ and $1e-3$, respectively. The input to the generation module was a concatenation of a retrieved idiom and a literal sentence together with token-level copy indicators. It was trained to generate idiomatic sentences.

\subsection{Human Evaluation Scale}

 For idiom inclusion, score $1$ denotes that the target phrase is not included in the input at all, $2$ denotes partial inclusion, and $3$ is for the complete inclusion. We report the average score over all samples for each baseline in each aspect.
 
\subsection{Results of Each Module in the Pipeline}

Here we provide evaluation of each module in the pipeline model individually. 

\subsubsection{Results of Idiom Retrieval}
Given a literal sentence, our pipeline first retrieves the idiom and then extracts the span to be replaced by the idiom. The retrieval module takes the idiom definition as well as the literal sentence as input to identify the idiom.
We evaluate the idiom retrieval module alone by reporting its accuracy of selecting the correct idioms as $0.714$. 
We also experimented with some variants. The details and results are in the Section \ref{sec:retrieval_app}. 

\subsubsection{Results of Span Extraction}

In the pipeline, the span extractor used  the idiom definition. Its F1 score in span extraction is $0.706$. Similar to the evaluation of the retrieval module, we also explored how the input and the module order affected the performance of the span extractor.  The details and results are in the Section \ref{sec:retrieval_app}. The setting of \textit{retrieve-then-extract with idiom definition} gives the best performance in both retrieval and span extraction.

\subsubsection{Results of Generation}
\begin{table}[t]
\small
    \centering
    \begin{tabular}{m{2.5cm}|c|c}
    \thickhline
        Accuracy & Guided \textsc{CopyNet} & \textsc{CopyNet} \\
        \thickhline
        Idiom part & 0.603 & 0.532 \\
        \hline
        Non-idiom part & 0.854 & 0.792 \\
        \thickhline
    \end{tabular}
    \caption{Accuracy of idiom and non-idiom parts in outputs.} 
    \label{tab:copy_acc}
\end{table}

We show in Table \ref{tab:overall2} that  Guided \textsc{CopyNet} outperforms  original \textsc{CopyNet} in terms of the quality of the generated idiomatic sentences by using automated metrics. 
The outputs consist of two main parts: the non-idiom words from the literal sentence and the retrieved idiom. Besides evaluating the complete sentences, we also compared the outputs of the two generators in terms of their accuracy in capturing the idiom  and the non-idiom parts respectively (shown in Table \ref{tab:copy_acc}). The accuracy of the idiom part denotes the percentage of words in the idioms captured in the output on an average. The accuracy of non-idioms is the percentage of the non-idiom words from the literal sentences retained in the outputs. We see that Guided \textsc{CopyNet} performs better than  original \textsc{CopyNet} in both retaining the non-idiom parts and incorporating the idiomatic words. 

As can be seen in Table \ref{tab:overall2}, the pipeline with Guided \textsc{CopyNet} is better than the pipeline with the rule-based generator.

For each idiom category (by lexical rigidity), we report the performance of both methods in Table~\ref{tab:fixedness}. We see that the rule based method fares slightly better than Guided \textsc{CopyNet} for fixed idioms. This is because the fixed idioms remain invariant regardless of the context and simple replacement done by the rule based method was sufficient for this category. As for the other idiom categories which require syntactic changes, Guided \textsc{CopyNet} outperforms the rule based method since it is able to modify the idioms  based on the context.

From the example in Table \ref{tab:examples}, we can see that  Guided \textsc{CopyNet} has the ability to change the tense of the corresponding verb ``mull'', while the rule based method does not. Other examples in Table \ref{tab:examples2} in the Appendix further show Guided \textsc{CopyNet}'s ability to generate more readable sentences by appropriately changing the verb tense, noun and pronoun forms.

\subsection{Variants for Idiom Retrieval and Span Extraction}
\label{sec:retrieval_app}

We  also report the results after exploring some variants of the retrieval module. Firstly, we replaced the idiom definition with the idiom in the input of the retrieval module. The corresponding retrieval accuracy is reported in the row \textit{retrieve-then-extract (use idiom)}. As shown in Table \ref{tab:acc12}, the accuracy drops from $0.714$ to $0.482$ when the definition is replaced with idiom in the input. The reason of the degraded retrieval performance is that many idioms are non-compositional, i.e., the idiomatic meaning cannot be inferred from their component words. This confirms that the definitions are more useful in conveying their meaning than the idioms themselves. 

The other variant we experimented with was to switch the retrieval and extraction module, i.e., we first identified the span to be removed, and then retrieved the idiom that should replace the extracted span. In Table \ref{tab:acc12}, its retrieval accuracy is reported in the rows of \textit{extract-then-retrieve}. We again report the performance of the retrieval with the definition and with the idiom respectively. It is shown that the setting of \textit{retrieve-then-extract} achieves better retrieval accuracy than the setting of \textit{extract-then-retrieve}.

For span extraction, when we replace the definition with the idiom in its input, the F1 scores decreased by $17.3\%$. When the order of retrieval module and span extractor was changed, the F1 score decreased by $30.3\%$ as shown in the Table \ref{tab:acc12}.

\begin{table}[t]
\small
    \centering
    \begin{tabular}{c|c|c}
    \thickhline
        Module Order & Accuracy & F1 \\
        \thickhline
        Retrieve-then-extract (use definition) & 0.714 & 0.706 \\
        \hline
        Extract-then-retrieve (use definition) & 0.291 & 0.403\\
        \hline
        Retrieve-then-extract (use idiom) & 0.482 & 0.533 \\
        \hline
        Extract-then-retrieve (use idiom) & 0.224 & 0.416 \\
        \thickhline
    \end{tabular}
    \caption{Accuracy for idiom retrieval and F1 score for span extraction.}
    \label{tab:acc12}
\end{table}

\subsection{More Examples of Generated Outputs}

Table \ref{tab:examples2} gives some more examples of literal sentences, the reference outputs, and idiomatic sentences generated by different models.

\begin{table*}[t]
\small
    \centering
    \begin{tabular}{c m{10cm}}
    \thickhline
    literal sentence & Don't {\color{red}{just imagine having a lot of money}}, just find some work to earn money. \\
    \hline
    Reference & Don't {\color{blue}{build castles in the air}}, just find some work to earn money. \\
    \hline
    Seq2Seq & Don't {\color{red}{just imagine having a lot of}} business, just find some work to earn money. \\
    Transformer & don't {\color{red}{just imagine having a lot of money}} just and find some work to earn money does to earn money . \\
    Seq2Seq with copy & don't build unk in a bad situation just find some work to earn money \\
    Transformer with copy & don't {\color{red}{just imagine having}} a killing just find some work to earn money . \\
    Ours - original \textsc{CopyNet} & Don't bread bread and just just and find some work to earn earlier. \\
    Ours -Rule based & Don't just {\color{green}{bread and butter}}, just find some work to earn money. \\
    Ours - Guided \textsc{CopyNet} & Don't just {\color{green}{bread and butter}} just find some work to earn money now. \\
    \hline
    \hline
    literal sentence & They promised that they will {\color{red}{search in all possible places}} to find the solution to the problem. \\
    \hline
    Reference & They promised that they will {\color{blue}{leave no stone unturned}} to find the solution to the problem. \\
    \hline
    Seq2Seq & They promised that they will {\color{red}{search in all possible places}} to find the solution to the problem. \\
    Transformer & they promised that they will {\color{red}{search in all possible places}} to find the solution to the problem . \\
    Seq2Seq with copy & they promised that they will leave in a bad situation worse than to the the to to to to to the \\
    Transformer with copy & they promised that they will {\color{blue}{leave no}} {\color{red}{search in all possible places}} to find the solution to the problem . \\
    Ours - original \textsc{CopyNet} & They promised that they will {\color{blue}{leave no stone unturned}} find find the the \\
    Ours -Rule based & they promised that they will {\color{blue}{leave no stone unturned}} to find the solution to the problem \\
    Ours - Guided \textsc{CopyNet} & They promised that they will {\color{blue}{leave no stone unturned}} to find the solution to the problem \\
    \hline
    \hline
    literal sentence & I am starting a {\color{red}{precision strike}} against my socks and vow to find the pair for each one of them in this messy drawer. \\
    \hline
    Reference & I am starting a {\color{blue}{surgical strike}} against my socks and vow to find the pair for each one of them in this messy drawer. \\
    \hline
    Seq2Seq & I am starting a blow-by-blow rise against my socks and gesturing to find the pair for each one of them. \\
    Transformer & i am starting a {\color{red}{precision strike}} against my socks and vow to find the pair for each one of them in this can drawer . \\
    Seq2Seq with copy & i am starting a unk strike against my wishes and unk to find the pair for each one of them in this unk drawer \\
    Transformer with copy & i am starting a breakout strike against my socks and vow to find the pair for each one of them in this messy drawer . \\
    Ours - original \textsc{CopyNet} & I am starting a strike strike against my socks and vow to find the pair for each one of them in this messy \\
    Ours -Rule based & i am starting a {\color{blue}{surgical strike}} strike against my socks and vow to find the pair for each one of them in this messy drawer \\
    Ours - Guided \textsc{CopyNet} & I am starting a {\color{blue}{surgical strike}} against my socks and vow to find the pair for each one of them in this messy drawer. \\
    \hline
    \hline
    literal sentence & She has started {\color{red}{reasoning out}} her career now and I would like to let her try things out. \\
    \hline
    Reference & She has started {\color{blue}{making sense}} about her career now and I would like to let her try things out. \\
    \hline
    Seq2Seq & She has started hot and her career now and I would like to let her try and out. \\
    Transformer & eos example \\
    Seq2Seq with copy & she has started  {\color{blue}{making sense}} of her career now and i would like to let her try things \\
    Transformer with copy & she has started {\color{red}{reasoning out}} her career now and i would like to let her go out with things out . \\
    Ours - original \textsc{CopyNet} & She has started out her career now I and I would like to let her try things things \\
    Ours -Rule based & she has started {\color{blue}{make sense}} out her career now and i would like to let her try things out \\
    Ours - Guided \textsc{CopyNet} & She has started {\color{blue}{making sense}} about her career now and I would like to let her try things out \\
    \thickhline
    \end{tabular}
    \caption{Samples of generated idiomatic sentences. Text in {\color{red}{red}} represents the literal phrases in the intput sentences. Text in {\color{blue}{blue}} represents the appropriate idioms included in the outputs. Text in {\color{green}{green}} represents the idioms that are poorly retrieved and included in the outputs.}
    \label{tab:examples2}
\end{table*}


\end{document}